\documentclass[preprint,12pt,author,numbers,sort&compress]{elsarticle}

\pdfoutput=1
\usepackage{amssymb}
\usepackage{multirow}
\usepackage{threeparttable}
\usepackage{graphicx}
\usepackage{booktabs}
\usepackage{tabularx}
\usepackage{array}
\usepackage{amsmath}
\usepackage{lscape}
\usepackage{longtable}
\usepackage{epstopdf}
\usepackage{graphicx}    
\usepackage{booktabs}    
\usepackage{colortbl}    
\usepackage{threeparttable}
\usepackage{rotating}
\usepackage[nodots]{numcompress}
\usepackage{natbib}

\newcommand{\PreserveBackslash}[1]{\let\temp=\\#1\let\\=\temp}
\newcolumntype{C}[1]{>{\PreserveBackslash\centering}p{#1}}
\newcolumntype{R}[1]{>{\PreserveBackslash\raggedleft}p{#1}}
\newcolumntype{L}[1]{>{\PreserveBackslash\raggedright}p{#1}}
\newtheorem{definition}{Definition}[section]
\newtheorem{lemma}{Lemma}[section]

\journal{APPLIED MATHEMATICS AND COMPUTATION}

\begin{document}

\begin{frontmatter}



\title{Uncertainty measurement with belief entropy on interference effect in Quantum-Like Bayesian Networks}


\author[address1]{Zhiming Huang}
\author[address2]{Lin Yang}
\author[address1]{Wen Jiang\corref{label1}}

\cortext[label1]{Corresponding author: Wen Jiang, School of Electronics and Information, Northwestern  Polytechnical University, Xi'an, Shannxi 710072, China. Tel: +862988431267. E-mail: jiangwen@nwpu.edu.cn.}

\address[address1]{School of Electronic and Information, Northwestern Polytechnical University, Xi'an 710072, China}
\address[address2]{China Equipment System Engineering Company, Beijing, 100039}

\begin{abstract}
Social dilemmas have been regarded as the essence of evolution game theory, in which the prisoner's dilemma game is the most famous metaphor for the problem of cooperation. Recent findings revealed people's behavior violated the Sure Thing Principle in such games. Classic probability methodologies have difficulty explaining the underlying mechanisms of people's behavior.
In this paper, a novel quantum-like Bayesian Network was proposed to accommodate the paradoxical phenomenon. The special network can take interference into consideration, which is likely to be an efficient way to describe the underlying mechanism. With the assistance of belief entropy, named as Deng entropy, the paper proposes Belief Distance to render the model practical. Tested with empirical data, the proposed model is proved to be predictable and effective.
\end{abstract}


\end{frontmatter}


\section{Introduction}\label{Introduction}
Prisoner's dilemma game is a famous metaphor for the problem of cooperation, which is a critical issue in evolutionary game theory \cite{Wang2017Onymity,Deng2017Sensors922}. If two players all defect, the payoff will be lower than if they all cooperates, as shown in Table \ref{payoff}. The paradoxical findings are shown in Table \ref{experimentresultsinliterature}, where the unknown part is not equal to the last column. The violation of $The \ Sure \ thing \ Principle$ \cite{Zachow1978Positive} shows humans break the law of classic probability when making decision under risk \cite{Ellsberg1963Risk}. Many analytical mythologies have been made to the explanation of this phenomenon but the underlying mechanisms are still enigmatic. Nevertheless, the quantum theory seems to be a practical method to uncover the mystery lying behind this incredible phenomenon \cite{Pawela2013Quantum,Bruza2009Is}.

    The quantum theory has been applied in many filed including information theory \cite{song2015finite}, decision making system \cite{yufirst2016,yu2015enhancing}, social and information networks \cite{wang2016statistical,yu2016system}. Busemeyer $et\ al.$ \cite{Busemeyer2009Empirical,Pothos2009A} proposed a Quantum Dynamical model based on a quantum version of a classical dynamical Markov model, which takes the process of making decisions into account of time evolution.  The quantum-like approach developed by Khrennikov \cite{Khrennikov2003Quantum} is based on contextual probabilities which can be applied to many domains like cognitive science economics, game theory, etc \cite{Khrennikov1997Non,Khrennikov2012On,Khrennikov2009Quantum}. Masanari $et\ al.$ \cite{Asano2011Quantum,Asano2012Quantum} proposed a quantum-like model to simulate the brain function.Li $et\ al.$ \cite{Li2011Quantum} proposed a quantum strategies into evolutionary games.

  Though there are many models based on quantum probability theory, few of them are predictable.  Inspired by the work \citep{Moreira2016Quantum,Iqbal2010Constructing,Shah2008Heuristics}, we propose a novel Bayesian Networks model based on quantum probability. This paper does not consider the noise effect in the quantum information systems \cite{Situ2016Relativistic,Situ2016Noise}. In this model, the violation of rational decision making in many experiments like Prisoner's dilemma game and the Two Stage Gambling game is characterized as interference effect between competing states. This paper regards man's mental beliefs as wave functions. Before the final decision is made, all potential decisions coexists in man's mind. Such uncertainty is like superposition state of wave functions \cite{Moreira2014Interference}. The interference effect is actually influenced by the partiality of the man towards to the decisions. Once the interference effect is determined, the man's behavior can be predicted and described by quantum probability theory. This paper proposes Belief Distance to measure the uncertainty with the assistance of belief entropy, named as Deng entropy. Uncertainty processing in decision making was firstly developed by Michèle and Jean Yves \cite{Cohen1985Decision} and the uncertainty can be measured based on distance \cite{AnImprovedAPIN2017,mo2016generalized}.   The knowledge to the uncertainty in decision making can help psychologists predict the behavior of humans with few fit errors. With the ability to compute the uncertainty of decision, the proposed model is predictable and simple for calculating.
  \begin{table}[]
\centering
\caption{Payoffs table}
\label{payoff}
\begin{tabular}{@{}lll@{}}
\toprule
A/B & 0   & 1   \\ \midrule
0   & 4/4 & 2/5 \\
1   & 5/2 & 3/3 \\ \bottomrule
\end{tabular}
\end{table}
\section{Organization of this paper}
 This paper is organized in the following manner. In section \ref{Preliminaries}, basic mathematical preliminaries will be introduced. In this section, a kind of belief entropy, called Deng entropy, will be introduced, which plays an important role in the model. After that the Bayesian model based on quantum probability will be presented in section \ref{proposedmodel}. Numerical examples will be illustrated in section \label{Numericalexample} to show how this model works. In the end, the proposed model will be compared with two models proposed in other literature to show its effectiveness.
\begin{table}[]
\tiny
\begin{threeparttable}
\centering
\caption{Experiment results of Prisoner's dilemma game from literature\citep{Moreira2016Quantum} }
\label{experimentresultsinliterature}
\begin{tabular*}{\columnwidth}{@{\extracolsep{\fill}}@{~~}ccccc@{~~}}
\toprule
Literature                  & \multicolumn{1}{l}{Known to Defect} & \multicolumn{1}{l}{Known to Collaborate} & \multicolumn{1}{l}{Unknown} & \multicolumn{1}{l}{Classical Probability} \\ \midrule
Shafir and Tversky, 1992    & 0.9700                              & 0.8400                                   & 0.6300                      & 0.9050                                    \\
Li and Taplin,2002          & 0.8200                              & 0.7700                                   & 0.7200                      & 0.7950                                    \\
Busemeyer et al., 2006a     & 0.9100                              & 0.8400                                   & 0.6600                      & 0.8750                                    \\
Hristova and Grinberg, 2008 & 0.9700                              & 0.9300                                   & 0.8800                      & 0.9500                                    \\ \midrule
Average                     & 0.8700                              & 0.7400                                   & 0.6400                      & 0.8050                                    \\ \bottomrule
\end{tabular*}
\begin{tablenotes}
\item[1]  The second column (Known to Defect) means the probability of the second player choose to betray when he/she knows the first player has chosen to betray. The third column(Known to Collaborate) means the probability of the second player choose to betray when he/she knows the first player has chosen to cooperate. The fourth column (Unknown) means the probability of the second player choose to betray without any information about the first player's action. The final column(Classical probability) means the probability calculated by the classic probability theory.
\end{tablenotes}
\end{threeparttable}
\end{table}

\section{Preliminaries}\label{Preliminaries}

\subsection{Belief Entropy}

Many contributions \cite{Vourdas2015Mobius,Vourdas2014Lower,Vourdas2014Quantum} have been made to interpret quantum probability into Dempster-Shafer probability, in which basic belief assignment is used to describe the probability of an event\cite{Zhang2016ANP,Tang2017A}. A new belief entropy, named as Deng entropy \cite{dengentropy} is a measure of uncertainty of basic belief assignment \cite{deng2015Generalized,Jiang2017mGCR}. Basic belief assignment(BBA) is widely used in the field of information fusion \cite{Jiang2017Ordered,Jiang2016CAIE,Wangjw2016evidence} which has been applied in many fields like Failure Mode and Effect Analysis \cite{jiang2017FMEA,jiang2017Ranking}, Fault Diagnose \cite{Yuan2016Modeling,Jiang2016sensor} and so on.


\begin{definition}
Let $\theta  = \left\{ {{H_1},{H_2},...,{H_N}} \right\}$ be a finite nonempty set of N elements which is mutually exclusive and exhaustive. Denote $P(\theta)$ as the power set composed of $2^N$ elements of $\theta$. The basic belief assignments(BBAs) function is defined as a mapping of the power set $P(\theta)$ to the value between $0$ and $1$.
$m: P(\theta) \to [0,1] $, which satisfies the following conditions:
\begin{equation}
\begin{aligned}
&m(\emptyset)=0 \\
&\sum\nolimits_{A \subseteq P(\theta )} {m(A) = 1}
\end{aligned}
\end{equation}
where the mass m(A) represents the support degree of evidence to event $A$.
\end{definition}


Shannon entropy, also named as information entropy, is the expected value of the information contained in each message which can be modeled by any flow of information.
\begin{definition}
The Shannon entropy is defined as follows:
\begin{equation}
H =  - \sum\nolimits_i {{P_i}} {\log _b}{P_i}
\end{equation}
where $P_i$ satisfies $\sum\nolimits_i {P_i} = 1$, $b$ is base of logarithm. When $b=2$, the unit of Shannon entropy is bit.
\end{definition}

The Belief entropy, named as Deng entropy, is introduced here to measure the uncertainty degree of BBAs, which is defined by:
\begin{definition}
\begin{equation}\label{DengEntropy}
{E_d} =  - \sum\nolimits_i {m(A)} \log \frac{{m(A)}}{{{2^{\left| A \right|}} - 1}}
\end{equation}
 Where $m$ is the BBAs function, and $A$ is the element of $P(\theta)$, $\left| A \right|$ is the cardinality of $A$. When $\left| A \right|$ is equal to $1$, the belief entropy will degenerate into Shannon entropy. The term $2^{\left| A \right|} - 1$ represents the potential states in A.
\end{definition}

\textbf{Example 1}
Assume there is a BBAs function m(a)=1. The Shannon entropy and Deng entropy are computed as follows:

$H=-1 \times log_2 1=0$

$E_d = -1\times log_2 \frac{1}{2^1 -1}=1$

This example shows if $\left| A \right|$ is equal to $1$, the belief entropy is similar with the classic Shannon entropy.

\textbf{Example 2}
 Given a set $\Theta=\{a,b,c \}$ with $m(\{a\})=\frac{1}{2}$ and $m(\{b,c\})=\frac{1}{2}$. The Deng entropy will be:

 $E_d = -\frac{1}{2} \times log_2 \frac{\frac{1}{2}}{2^2 - 1}$

The above examples show how Deng entropy works and overcomes the insufficiency of Shannon entropy  when measuring the uncertainty in problems like Example 2.
\subsection{The Classic Bayesian Network and the Quantum-like Bayesian Model}
\subsubsection{Classic Bayesian Network}

\begin{figure}
\centering
\includegraphics[width=4in]{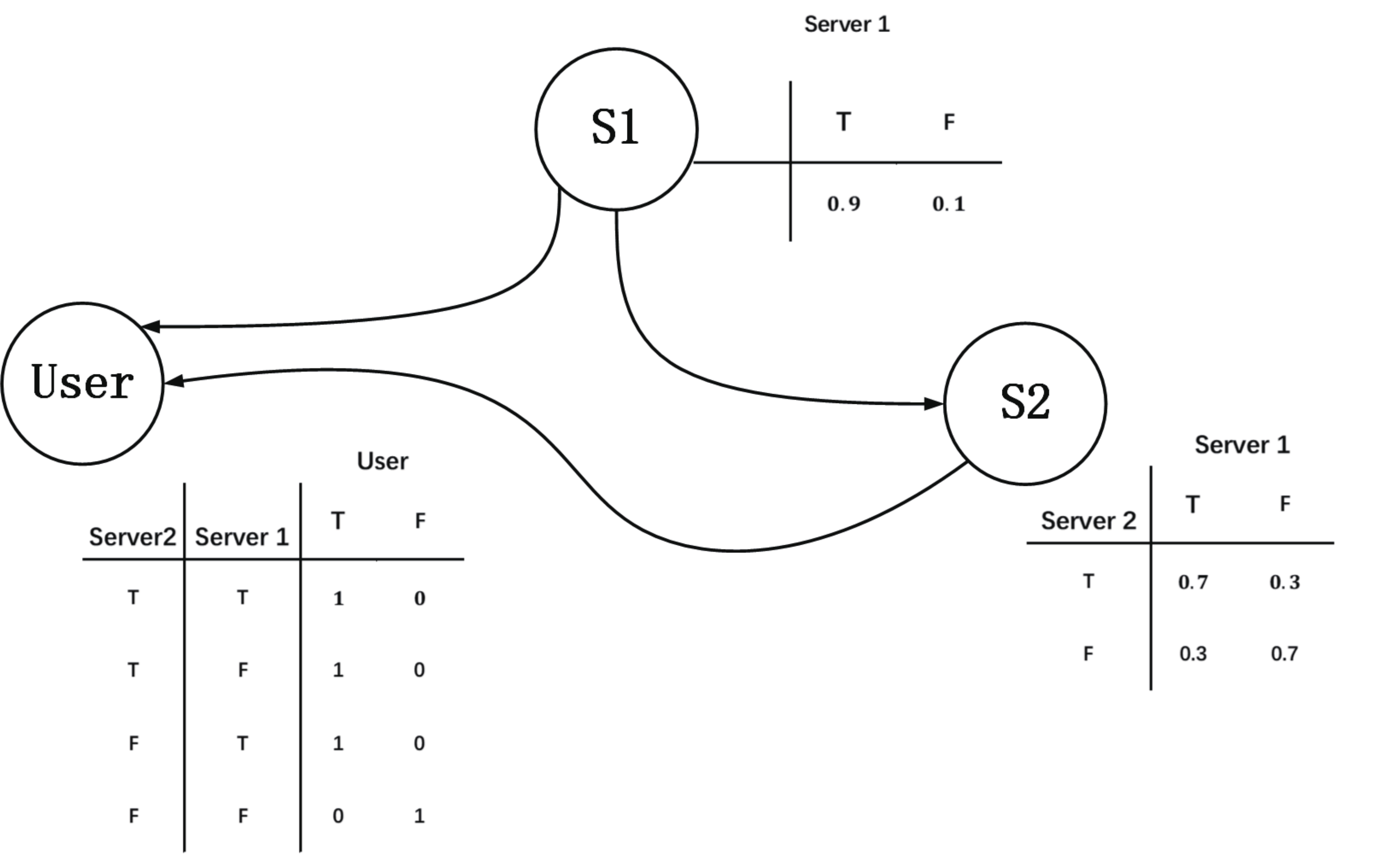}
\caption{An example of Bayesian Network}
\label{BayesianNetwork}
\end{figure}

A classic Bayesian Network is a kind of probabilistic directed acyclic graphical model, which has been successfully applied in the field of decision making \cite{Myung2005A}. In this model, a set of random variables and their conditional dependencies are represented via a directed acyclic graph. Each node that represents a variable is associated with a conditional probability table, as shown in Fig.\ref{BayesianNetwork}.
\begin{definition}
The full joint distribution of a Bayesian Network is defined by:
\begin{equation}\label{classicfulljoint}
\Pr ({X_1},{X_2}...,{X_n}) = \prod\limits_{i = 1}^n {\Pr ({X_i}|Parents({X_i}))}
\end{equation}
\end{definition}
where $X$ is the list of variables, $Parents(X_i)$ means nodes pointing to $X_i$.
The model can answer any query with response of $yes$ or $no$ by using conditional probability formula and summing over all nuisance variables. For some query $X$, the inference is given by Eq.(\ref{123})
\begin{equation}\label{123}
\begin{split}
\Pr (X|e) = \alpha [\sum\limits_{y \in Y} {\Pr (X,e,y)} ] \\
where \quad \alpha  = \frac{1}{{\sum\nolimits_{x \in X} {{{\Pr }_c}(X = x,e)} }}
\end{split}
\end{equation}
where $e$ is the list of observed variables (nodes) and $y$ is the remaining unobserved variables(nodes) in the network, the $\alpha$ is the normalization factor for the distribution $\Pr (X|e)$ \cite{Russel2002Artificial}.

\textbf{Example:} Fig.\ref{BayesianNetwork} shows an example of Bayesian Network. Assume there are two servers $S1$ and $S2$ transmitting data packets to $User$. Apparently, the parent nodes of $User$ are $S1$ and $S2$ and the parent node of $S2$ is $S1$. Each node has a conditional probability table which represents if a packet is transmitted successfully. If there is a query, for example, what is the probability when $user$ successively receives one data packet. The inference is computed by Eq.(\ref{123}) as follows:

\begin{equation}
\begin{aligned}
&\Pr (One\ Packet)=\alpha \{Pr(S2=T|S1=F)*Pr(S1=F)\\
&+ Pr(S2=F|S1=T)*Pr(S1=T)\} = \alpha (0.3*0.1+ 0.3*0.9)= 0.3\alpha \\
&\Pr (two\ or\ zero\ Packets)=\alpha \{Pr(S2=T|S1=T)*Pr(S1=T)\\
&+ Pr(S2=F|S1=F)*Pr(S1=F)\} = \alpha (0.7*0.9+ 0.7*0.1)= 0.7\alpha \\
&\alpha = \frac {1} {\Pr (One\ Packet)+\Pr (two\ or\ zero\ Packets)} =1
\end{aligned}
\end{equation}
The above example shows the basic idea of Bayesian network and procedure of deriving inferences according to some queries.

\subsubsection{Quantum-like Bayesian Model}
Bayesian networks can split complex problem into small modules that can be combined to perform inferences \cite{Khrennikov2016Quantum,Barros2009Quantum}. The quantum-like Bayesian Model \cite{Moreira2016Quantum} replaces the real probability numbers in the classic probability Bayesian Network model with quantum probability amplitudes \cite{ROBERT1997QUANTUM,Leifer2008Quantum}.

The corresponding part of quantum-like Bayesian Network model to the application of Born's rule to Eq.(\ref{classicfulljoint}) is:
\begin{equation}\label{quantumfulljoint}
\Pr ({X_1},...,{X_n}) = |\prod\limits_{i = 1}^n {\psi ({X_i}|Parents({X_i}))} {|^2}
\end{equation}

The quantum application of Born's rule to the classic marginal probability distribution Eq.(\ref{123}) is defined by the equation below:
\begin{equation}\label{margin}
\begin{aligned}
&\Pr (X|e) = \partial  {\rm{|}}\sum\limits_Y {\prod\limits_x^N {\psi ({X_x}|Parents({X_x}),e,y)} } {{\rm{|}}^2}\\
\end{aligned}
\end{equation}
\begin{equation}\label{partial}
Where \quad \partial  = \frac{1}{{\sum\nolimits_{x \in X} {{{\Pr }_c}(X = x,e)} }} =1
\end{equation}

A quantum marginalization formula with interference effects \citep{Moreira2014Interference} emerges when the  Eq.(\ref{margin}) expands, as shown in below,

\begin{equation}\label{Interfernce}
\begin{aligned}
& \Pr (X|e) = \partial \sum\limits_{i = 1}^{|Y|} {|\prod\limits_x^N {\psi ({X_x}|Parents({X_x}),e,y = i)} {|^2}+ 2 \cdot Interference} \\
& Interference = \sum\limits_{i = 1}^{|Y| - 1} {\sum\limits_{j = i + 1}^{|Y|} {|\prod\limits_x^N {\psi ({X_x}|Parents({X_x}),e,y = i)} | \cdot }} \\
& \quad \quad \quad \quad \quad \quad \quad \quad {{|\prod\limits_x^N {\psi ({X_x}|Parents({X_x}),e,y = j)} |} \cdot \cos ({\theta _i} - {\theta _j})}\\
\end{aligned}
\end{equation}

\begin{figure}
\centering
\includegraphics[width=4in]{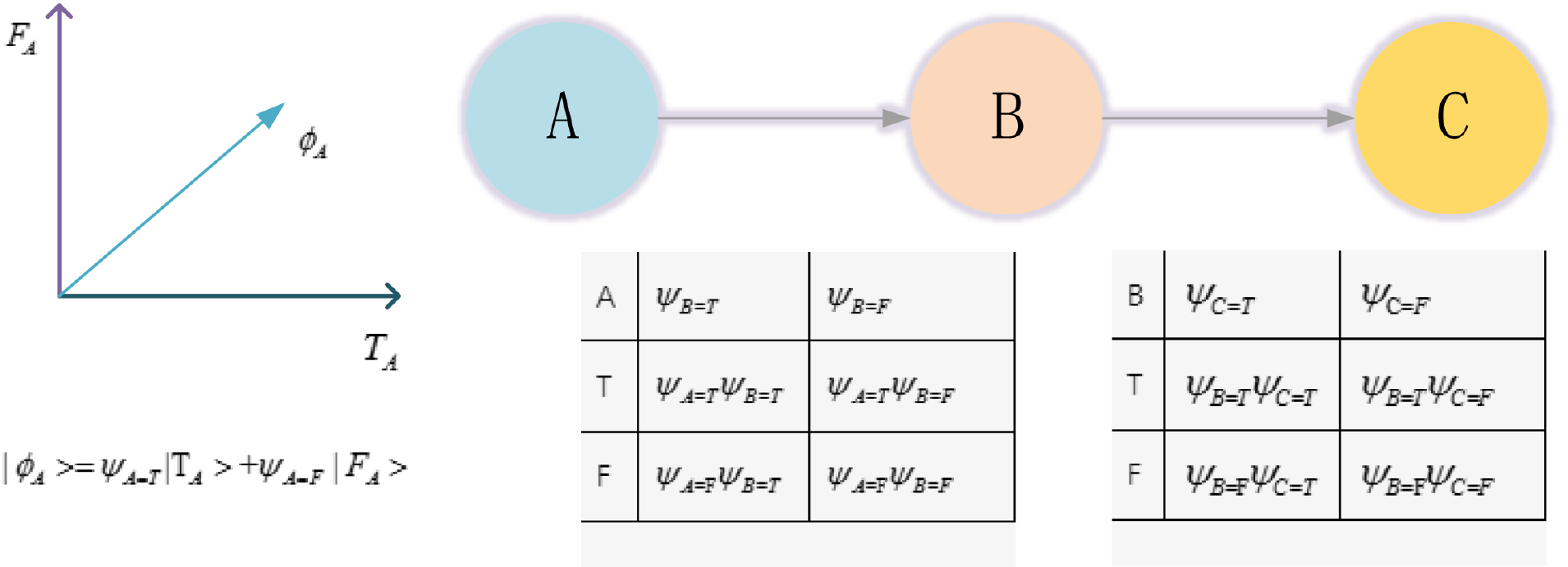}
\caption{An example of a Quantum-like Bayesian Network}
\label{model1}
\end{figure}

\textbf{Example:}  Fig.\ref{model1} shows an instance of Quantum-like Bayesian Network.

 This network can only answer queries with $yes$ or $no$ answer, which are regarded as base vectors $|0>$ and $|1>$. Fig.\ref{model2} shows any actions the node will take can be seen as wave functions characterized by base vectors $|0>$ and $|1>$, as defined by:
 \begin{equation}\label{basevecotrs}
 \begin{aligned}
 &|T>\ =\ cos\theta_T |1>\ +\ sin\theta_T |0>\ =\ e^{j\theta_T} \\
 &|F>\ =\ cos\theta_F |1>\ +\ sin\theta_F |0>\ =\ e^{j\theta_F}
 \end{aligned}
 \end{equation}

 Thus, the decision vector for node $A$ is defined by:
\begin{equation}
|{\phi _A} >\  =\ {\psi _{A = T}}{\rm{|}}{{\rm{T}}_A} > {\rm{\ +\ }}{\psi _{A = F}}|{F_A} >\ =\ \psi _{A = T}\cdot e^{j\theta_{T_A}}\ +\ \psi _{A = F}\cdot e^{j\theta_{F_A}}\\
\end{equation}
where the action states $|{F_A} >$ and $|{T_A} >$ means the actions the node can take. The index $A$ in $|{F_A} >$ and $|{T_A} >$ represents this decision is made by node $A$.

In the same way, Decision vector for node $B$ is
\begin{equation}\label{superposition}
|{\phi _B} >\  =\ {\psi _{B = T}}{\rm{|}}{{\rm{T}}_B} > {\rm{\ +\ }}{\psi _{B = F}}|{F_B} >\ =\ \psi _{B = T}\cdot e^{j\theta_{T_B}}\ +\ \psi _{B = F}\cdot e^{j\theta_{F_B}}\\
\end{equation}

For a query "what is the probability for B to adopt action T ?", the inference is computed by Eq.(\ref{margin}):
\begin{equation}\label{infer1}
\begin{aligned}
\Pr (T|A)\ &\ = \ \partial | \psi _{B = T}\cdot e^{j\theta_{T_B}}\cdot \psi _{A = T}\cdot e^{j\theta_{T_A}}\ +\ \psi _{B = T}\cdot e^{j\theta_{T_B}}\cdot \psi _{A = F}\cdot e^{j\theta_{F_A}}|^2\\
&\ =\ \partial | \psi _{B = T}\cdot \psi _{A = T}\cdot e^{j\theta_{1}}\ +\ \psi _{B = T}\cdot \psi _{A = F}\cdot e^{j\theta_{2}}|^2\\
&\ = \ \partial |\psi _{B = T}\cdot \psi _{A = T}\cdot e^{j\theta_{1}}\ +\ \psi _{B = T}\cdot \psi _{A = F}\cdot e^{j\theta_{2}}|\\
&\ \ \ \  \ \cdot |\psi _{B = T}\cdot \psi _{A = T}\cdot e^{j\theta_{1}}\ +\ \psi _{B = T}\cdot \psi _{A = F}\cdot e^{j\theta_{2}}|^*\\
&\ =\ \partial |\psi _{B = T}\cdot \psi _{A = T}|^2\ +\ |\psi _{B = T}\cdot \psi _{A = F}|^2\ \\
&\ \ \ \ + \ \psi _{B = T}\cdot \psi _{A = T}\cdot e^{j\theta_{1}} \cdot \psi _{B = T}\cdot \psi _{A = F}\cdot e^{-j\theta_{2}}\\
&\ \ \ \ + \ \psi _{B = T}\cdot \psi _{A = F}\cdot e^{j\theta_{2}} \cdot \psi _{B = T}\cdot \psi _{A = T}\cdot e^{-j\theta_{1}}\\
&\ =\ \partial |\psi _{B = T}\cdot \psi _{A = T}|^2\ +\ |\psi _{B = T}\cdot \psi _{A = F}|^2\ +\\
 &\ \ \ \ 2 \cdot |\psi _{B = T}\cdot \psi _{A = T} \cdot \psi _{B = T}\cdot \psi _{A = F}|\cdot cos(\theta_1 - \theta_2)
\end{aligned}
\end{equation}
\begin{equation}\label{infer2}
\begin{aligned}
\Pr (F|A)&\ = \ \partial |\psi _{B = F}\cdot \psi _{A = T}|^2\ +\ |\psi _{B = F}\cdot \psi _{A = F}|^2\ +\ \ \ \ \ \ \ \ \ \ \ \ \ \ \ \ \ \ \ \ \ \\
 &\ \ \ \ 2 \cdot |\psi _{B = F}\cdot \psi _{A = T} \cdot \psi _{B = F}\cdot \psi _{A = F}|\cdot cos(\theta_3 - \theta_4)
\end{aligned}
\end{equation}
where $\theta_1 = \theta_{T_B}+ \theta_{T_A}$ ,$\theta_2  = \theta_{T_B}+ \theta_{F_A}$, $\theta_3 = \theta_{F_B}+ \theta_{T_A}$, $\theta_4 = \theta_{F_B}+ \theta_{F_A}$.

Therefore, the answer to query is $\Pr (T|A)$ once the $\partial$ is determined by Eq.(\ref{partial}). This example illustrates the definition of Quantum-like Bayesian Network and the detail derivation of Eq.(\ref{Interfernce}).

\section{The proposed model}\label{proposedmodel}
Unlike the method in the literature \citep{Moreira2016Quantum}, this paper proposes a new way to calculate the interference value in the quantum-like Bayesian Network model. The biggest difference is this paper replaces the term $\cos (\theta_i - \theta_j)$ in Eq.(\ref{Interfernce}) with the uncertainty degree value $E_d$ calculated by Deng entropy. Deng entropy, also named as Belief entropy, is a powerful tool to measure the belief degree. The term $\cos (\theta_i - \theta_j)$ in Eq.(\ref{Interfernce}) is a degree of belief uncertainty in the quantum interference term. When prisoner has no information about the other prisoner, his/her decision is influenced by his/her belief about the rival's decision. That's why the classic probability framework can not describe the game properly because to some extant the prisoner is not totally "ignorant" about the other prisoner but has his/her own belief about the other part. This uncertain belief causes the interference term in the Bayesian Network model, which seems to be variable because the human's mind is changeable and is hard to measure. However, many experiments in literatures have revealed that the human's belief was inclined to certain degree, which means the interference term has a tendency value. Once the degree of  belief uncertainty could be measured, the model can be established to describe the behavior that violates the Sure Thing Principle. Here the Deng entropy is introduced to measure the belief degree and the results turns out to be fit for the model to describe the game.

\subsection{Acquisition of Belief Degree}

This paper presents such a concept that the existing of interference term is because the prisoner's belief to the other prisoner. According to the classic probability theory, as analysed in the above section, the probability of a prisoner to defect the other under unknown condition should be equal to $\frac{1}{2}(Pr (P2 = Defect|P1 = Cooperate)+Pr(P2 = Defect|P1 = Defect))$. But the experiment results in literature denied this, which means the interference term truly affects.  That's because actually the prisoner is not totally "ignorant" about the other, for he/she will predict the other prisoner's decision from his/her own perspective and then make self's decision. For every individual, every one has his/her own characteristics. When predicting other's decision from self's perspective, the result seems to be diverse. But it is known that there are something that is common for everyone called human nature which results in most people that they tend to have a same predication tendency.

\begin{definition}
  Belief Degree is defined by:
\begin{equation}
\begin{aligned}
&D_b = cos(\theta_i  - \theta_j)\\
\end{aligned}
\end{equation}
where $\theta_i$  and $\theta_j $ are angles in interference term in Eq.(\ref{Interfernce}). Belief degree represents people's predication toward their opponents and their belief tendency to certain actions in prisoners' dilemma game.
\end{definition}

This predication tendency or Belief Degree determines the value of interference term. According to the previous experiments shown in Table \ref{experimentresultsinliterature}, the value of interference term is inclined to a certain value, which means there indeed exists predication tendency or Belief Degree. Hence the Belief Degree can be determined as shown in following.

\begin{figure}
\centering
\includegraphics[width=3in]{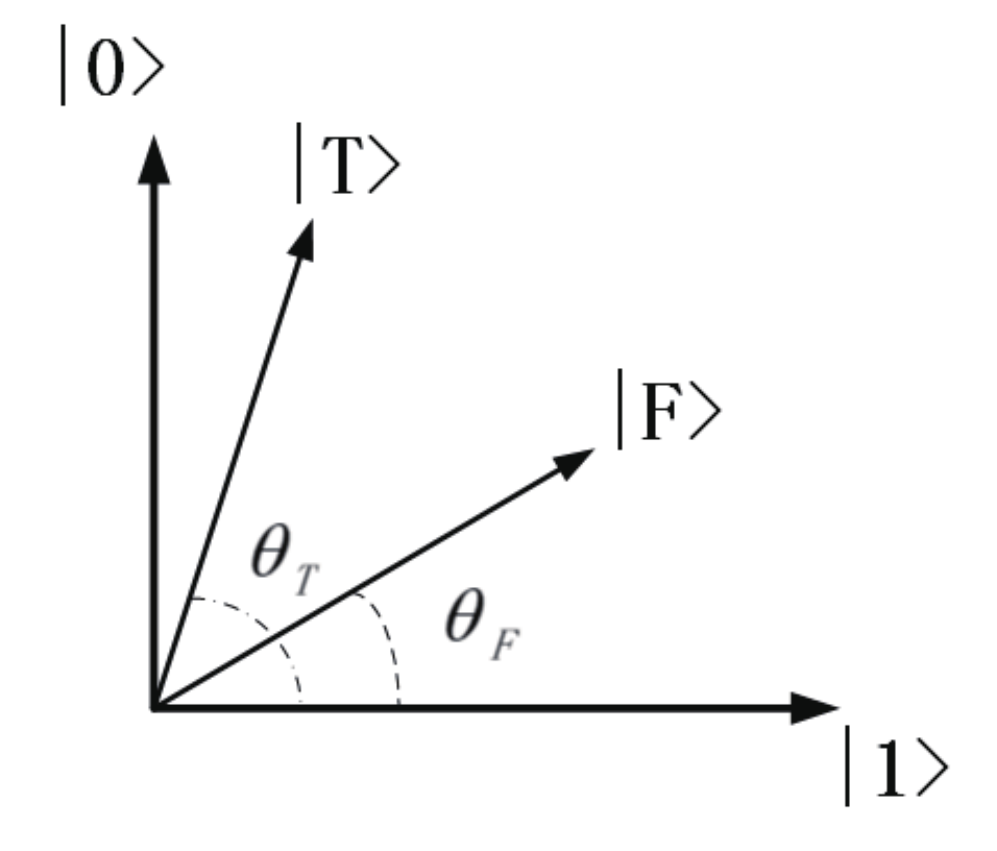}
\caption{vector representation}
\label{model2}
\end{figure}

\begin{figure}
\centering
\includegraphics[width=3in]{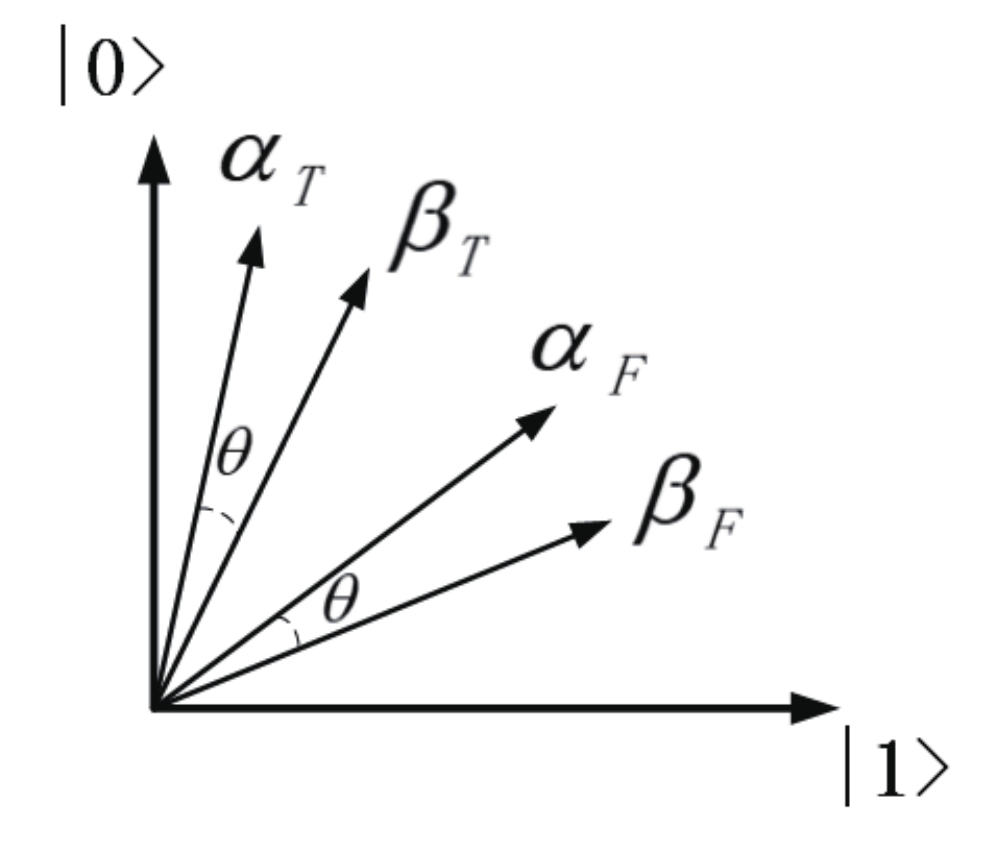}
\caption{vector representation of $\alpha$ and $\beta$}
\label{model6}
\end{figure}

The quantum marginalization formula comprises two parts, the classic probability term and interference term, as Eq.(\ref{Interfernce}) shows. It is the interference term that equips the model with ability to accommodate the violation of Sure Thing Principle. In this section, Deng entropy will be introduced to calculate the belief uncertainty to obtain the interference value.
Notice that the Eq.(\ref{basevecotrs}) has two basis states, as shown in Fig.\ref{model2}.
There always be two vectors representation of Eq.(\ref{Interfernce}), for the variable $X$ has two alternative value, $T$ and $F$.
\begin{equation}\label{vectors}
\left[ {\begin{array}{*{20}{c}}
{{\alpha _T}}\\
{{\beta _T}}
\end{array}} \right] = \left[ {\begin{array}{*{20}{c}}
{{\psi _{{P_N} = T}} \cdot {\psi _{{P_{Parents}} = T}}}\\
{{\psi _{{P_N} = T}} \cdot {\psi _{{P_{Parents}} = F}}}
\end{array}} \right]{\rm{ }}
\left[ {\begin{array}{*{20}{c}}
{{\alpha _F}}\\
{{\beta _F}}
\end{array}} \right] = \left[ {\begin{array}{*{20}{c}}
{{\psi _{{P_N} = F}} \cdot {\psi _{{P_{Parents}} = T}}}\\
{{\psi _{{P_N} = F}} \cdot {\psi _{{P_{Parents}} = F}}}
\end{array}} \right]
\end{equation}
Before applying Deng entropy to obtain the uncertain term $cos(\theta_i-\theta_j)$ in the interference term, we should process the data in Eq.(\ref{vectors}) firstly. The vector representation of Eq.(\ref{vectors}) is shown in Fig.\ref{model6}. As can be inferred from Eq.(\ref{infer1}) and Eq.(\ref{infer2}), two $\theta$ in the Fig.\ref{model6} have the same value. Though we have known the value of two pairs of $\alpha$ and $\beta$, the value of $\theta$ can hardly be determined through existing methods. One possible solution is just to regard $cos(\theta_i-\theta_j)$ as an uncertain variable, which can be replaced by belief degree $D_b$. Hence once the belief degree is determined, the interference term is settled. The belief degree can be determined through belief entropy, which can calculate the uncertainty from Belief Distance.

\begin{definition}
The Belief Distance is defined by:
\begin{equation}\label{heuristicdistance}
\begin{aligned}
&B_{d_X}=|{\alpha _X}{\rm{ + }}\frac{\alpha _X - \beta _X}{|\alpha _X + \beta _X - 1|}| \\
&where \quad |{\alpha _X} - 0.5| < |{\beta _X} - 0.5|
\end{aligned}
\end{equation}
\end{definition}
If $|{\alpha _X} - 0.5| \ge |{\beta _X} - 0.5|$, the position of $\alpha_X$ and $\beta_ X$ should be switched.

The Belief Distance measuring the deviation from $0.5$. If no information is provided, the value of $\alpha$ and $\beta$ would be $0.5$ because node A has two actions with each amplitude $\sqrt{5}$ and so does node B. $\frac{|\alpha _X - \beta _X|}{|\alpha _X + \beta _X - 1|}$ is actually a derivation of $\frac{|\alpha _X -0.5|-|\beta _X -0.5|}{|\alpha _X -0.5|+|\beta _X -0.5|}$ .
\begin{lemma}
With the relative deviation information provided, Belief degree can be computed by Eq.(\ref{heuristicdistance}) and Eq.(\ref{DengEntropy}):
\begin{equation}\label{Heuristicwithdengentropy}
D_b=-{E_d} =  \sum\nolimits_x {{{\rm{B}}_{{d_x}}}} \log \frac{{{{\rm{B}}_{{d_x}}}}}{{{2^{\left| {{A_i}} \right|}} - 1}}
\end{equation}
${{\left| {{A_i}} \right|}}$ means the number of unobserved variables.
\end{lemma}
In quantum mechanics, the $cos(\theta_1 - \theta_2)$ is given by the inner product between two wave functions\cite{Busemeyer2012Hierarchical}, which describes the subtraction of phases of the two wave function. Because it is difficult to compute $cos(\theta_i-\theta_j)$ from geometric perspective, this paper just regards it as a variable which can be computed through belief entropy.

\section{Numerical example}\label{Numericalexample}
In this section, the proposed method will be applied in the Bayesian Network model to analyze the average results presented in Table \ref{experimentresultsinliterature}.  The process could be summarized as below.

\begin{figure}
\centering
\includegraphics[width=5in]{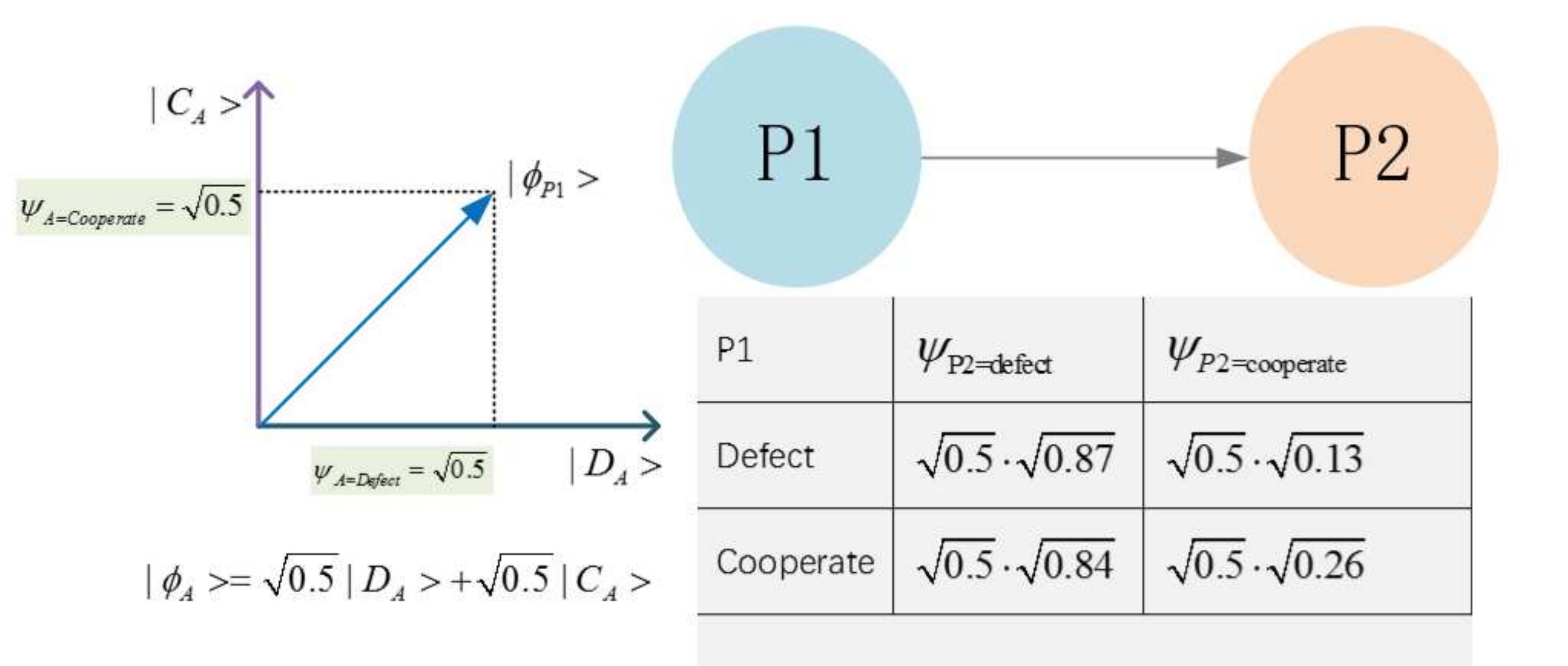}
\caption{Bayesian Network model for the Prisoners' dilemma game with the average results from Table \ref{experimentresultsinliterature}}
\label{model3}
\end{figure}

\textbf{Step 1: Create the model for the problem:}
If nothing is told, the first participant in the Prisoner's dilemma game will choose $Defect$ or $Cooperate$ with probability of $0.5$. The reason we assume the probability equals to $0.5$ is that the first participant in the model do not have $parents$ and nothing is told to him/her. However, in the real situation, the participants will wonder the other participant's action and make decisions based on the judgement. Therefore the assumption that the probability $0.5$ is uncertain. In the Eq.(\ref{Heuristicwithdengentropy}), ${{\left| {{A_i}} \right|}}$ means the number of variables whose decision are not sure. Under this situation, the first participant's decision assumed by us is not exactly certain, so the term ${{\left| {{A_i}} \right|}}$ will equal to $1$. With the data from Table \ref{experimentresultsinliterature}, we can establish a model as shown in Fig.\ref{model3};

\textbf{Step 2: Compute the Belief distance:}
According to Fig.\ref{model3}, the Eq.(\ref{vectors}) can be paraphrased as below:
\begin{equation}
\begin{array}{l}
\left[ {\begin{array}{*{20}{c}}
{{\alpha _T}}\\
{{\beta _T}}
\end{array}} \right] = \left[ {\begin{array}{*{20}{c}}
{{\psi _{{P_N} = T}} \cdot {\psi _{{P_{Parents}} = T}}}\\
{{\psi _{{P_N} = T}} \cdot {\psi _{{P_{Parents}} = F}}}
\end{array}} \right]{\rm{ = }}\left[ {\begin{array}{*{20}{c}}
{\sqrt {0.5}  \cdot \sqrt {0.26} }\\
{\sqrt {0.5}  \cdot \sqrt {0.13} }
\end{array}} \right] = \left[ {\begin{array}{*{20}{c}}
{0.3606}\\
{0.2550}
\end{array}} \right]\\
\left[ {\begin{array}{*{20}{c}}
{{\alpha _F}}\\
{{\beta _F}}
\end{array}} \right] = \left[ {\begin{array}{*{20}{c}}
{{\psi _{{P_N} = F}} \cdot {\psi _{{P_{Parents}} = T}}}\\
{{\psi _{{P_N} = F}} \cdot {\psi _{{P_{Parents}} = F}}}
\end{array}} \right] = \left[ {\begin{array}{*{20}{c}}
{\sqrt {0.5}  \cdot \sqrt {0.74} }\\
{\sqrt {0.5}  \cdot \sqrt {0.87} }
\end{array}} \right] = \left[ {\begin{array}{*{20}{c}}
{0.6083}\\
{0.6595}
\end{array}} \right]
\end{array}
\end{equation}
In this way, one can calculate the belief distance with Eq.(\ref{heuristicdistance}). Here we take the calculation process of ${\alpha _F}$ and ${\beta _F}$ for example:
notice that $|{\beta _T}-0.5| > |{\alpha _T}-0.5|$. The Belief Distance for ${\alpha _F}$ and ${\beta _F}$ is:
\begin{equation}
{B_{{d_T}}} = |0.6083{\rm{ + }}\frac{0.6083 - 0.6595}{|0.6083 + 0.6595 - 1|}| = 0.41711
\end{equation}
And the Belief distance for ${\alpha _T}$ and ${\beta _T}$ can be computed in the same way:
\begin{equation}
{B_{{d_F}}} = |0.3606{\rm{ + }}\frac{0.3606 - 0.2550}{|0.3606 + 0.2550 - 1|}| = 0.63531
\end{equation}

\textbf{Step 3: Calculate the belief degree using Deng entropy:}
In the Step 2 we obtain the Belief Distance $B_{{d_T}}$ and $B_{{d_F}}$. The Belief Distance represents the inner connection between two actions decided by the participants, as a reflection of a prisoner's belief to the other. The Deng entropy is an efficient tool to reveal this connection. In Step 1, we have analyzed that the term $|A_i|$ in the Eq.(\ref{Heuristicwithdengentropy}) equals to 1. Hence the results of Eq.(\ref{Heuristicwithdengentropy}) is:
\begin{equation}
D_b=-{E_d} =  0.41711 \cdot \log \frac{{0.41711}}{{{2^1} - 1}} + {\rm{0}}{\rm{.63531}} \cdot \log \frac{{{\rm{0}}{\rm{.63531}}}}{{{2^1} - 1}} = - 0.9420
\end{equation}
The $D_b$ will replace the term $cos(\theta_i-\theta_j)$ in the Eq.(\ref{Interfernce}) to perform the probabilistic interference.
\begin{equation}
\begin{aligned}
&\Pr (P2=Defect) = \partial [|{\psi _{P2 = D|P1 = D}}{|^2} + |{\psi _{P2 = D|P1 = C}}{|^2} \\
&+ 2 \cdot |{\psi _{P2 = D|P1 = D}}| \cdot |{\psi _{P2 = D|P1 = C}}| \cdot \cos (\theta_1-\theta_2)]\\
&= \partial [0.5 \times 0.87 + 0.5\times 0.74 + 2 \cdot \sqrt {0.5 \times 0.87}  \cdot \sqrt {0.5 \times 0.74}  \cdot  -0.9420 ]
\end{aligned}
\end{equation}
$\Pr (P2=Cooperate)$ can be obtained i the same way:
\begin{equation}
\begin{aligned}
&\Pr (P2=Defect)=\partial 0.04917 \\
&\Pr (P2=Cooperate)=\partial 0.02182
\end{aligned}
\end{equation}

And the final result is:
\begin{equation}
\begin{aligned}
&\Pr (P2=Defect)= \frac{\partial 0.04917}{\partial 0.04917+\partial 0.02182}=0.6926 \\
&\Pr (P2=Defect)= \frac{\partial 0.02182}{\partial 0.04917+\partial 0.02182}=0.3074
\end{aligned}
\end{equation}
Compare the result with probability in Table \ref{experimentresultsinliterature}, the model this paper proposes produces a result with fit error percentage of ${8.2{\%}}$.

\begin{figure}
\centering
\includegraphics[width=5in]{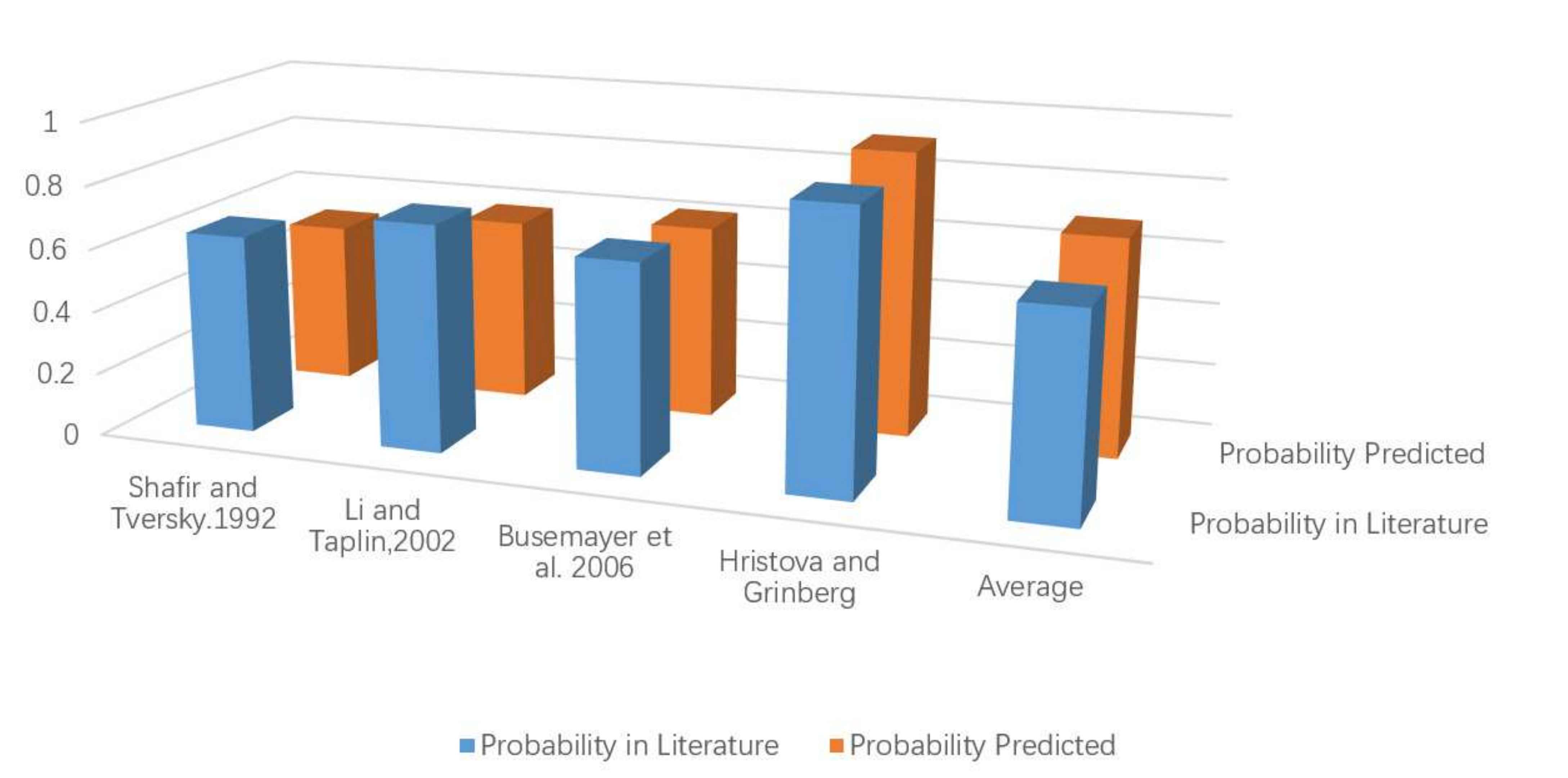}
\caption{Comparison of results from the literature and results predicted by the model}
\label{model4}
\end{figure}

Fig.\ref{model4} shows the comparison of results from literature and prediction of model, from which we can see that the model prediction is coincident to the probability observed with little fit errors.

\section{Conclusion}\label{conclusion}
  Quantum Bayesian Network inherits inference ability from classic Bayesian network and has the ability to explain the violation of $Sure\ Thing\ Principle$. The model proposed by this paper successfully described the paradoxical phenomenon in Prisoners' dilemma game. Unlike other existing methods, the proposed model regards the violation as an effect of interference and utilizes the concept of "Belief Degree" to make prediction though belief entropy.
  the model is compared with two other Quantum models. The first model (model 1) is the Quantum Prospect Decision Theory(QPDT) model developed by Yukalov and Sornette\cite{Yukalov2015Quantum,Yukalov2011Decision,Yukalov2010Entanglement,VYACHESLAV2010MATHEMATICAL}. In (QPDT) model, a static heuristic is used to predict the results. The second model is the Quantum-Like Bayesian Network proposed by Moreira\cite{Moreira2016Quantum}, in which a dynamic heuristic is used to predict the results.
\begin{table}[]
\centering
\caption{Comparison between proposed model and other two models in literature}
\label{comparisonmodel}
\resizebox{\textwidth}{!}{%
\begin{threeparttable}
\begin{tabular}{@{}cccccccc@{}}
\toprule
\textbf{Literature}         & \textbf{\begin{tabular}[c]{@{}c@{}}Pr(Defect)\\ Observed\end{tabular}} & \textbf{\begin{tabular}[c]{@{}c@{}}Pr(Defect)\\ Computed(Model 1)\end{tabular}} & \textbf{\begin{tabular}[c]{@{}c@{}}Fit errors\\ model 1\end{tabular}} & \textbf{\begin{tabular}[c]{@{}c@{}}Pr(Defect)\\ Computed(model 2)\end{tabular}} & \textbf{\begin{tabular}[c]{@{}c@{}}Fit errors\\ model 2\end{tabular}} & \textbf{\begin{tabular}[c]{@{}c@{}}Pr(Defect)\\ Computed(Proposed model)\end{tabular}} & \multicolumn{1}{l}{\textbf{\begin{tabular}[c]{@{}l@{}}Fie errors\\ Proposed model\end{tabular}}} \\ \midrule
Li and Taplin,2002 1        & 0.8667                                                                 & 0.6334                                                                          & 0.2692                                                               & 0.8113                                                                          & 0.0639                                                               & 0.8623                                                                                 & 0.0051                                                                                   \\
Li and Taplin,2002 2        & 0.7000                                                                 & 0.5333                                                                          & 0.2381                                                               & 0.7006                                                                          & 0.0009                                                               & 0.6691                                                                                 & 0.0441                                                                                   \\
Li and Taplin,2002 3        & 0.7667                                                                 & 0.5500                                                                          & 0.2826                                                               & 0.7159                                                                          & 0.0663                                                               & 0.7005                                                                                 & 0.0863                                                                                   \\
Busemeyer et al.,2006a      & 0.6600                                                                 & 0.6250                                                                          & 0.0531                                                               & 0.7995                                                                          & 0.2113                                                               & 0.6069                                                                                 & 0.0805                                                                                   \\
Hristova and Grinberg, 2008 & 0.8800                                                                 & 0.7000                                                                          & 0.2045                                                               & 0.8968                                                                          & 0.0191                                                               & 0.9045                                                                                 & 0.0279                                                                                   \\ \midrule
Average fit error           & -                                                                      & -                                                                               & \textbf{0.2095}                                                      & -                                                                               & \textbf{0.0723}                                                      & -                                                                                      & \textbf{0.04878}                                                                         \\ \bottomrule
\end{tabular}
\begin{tablenotes}
\item[Li and Taplin,2002 1] we use 3 experiments from literature\cite{Li2002Examining} and number them with 1,2 and 3 after the authors' name.
\end{tablenotes}
\end{threeparttable}%
}
\end{table}
From Table \ref{comparisonmodel} we can clearly notice that the average fit errors of proposed model is smaller than other two models, which shows the new method for Quantum Bayesian Network proposed by this paper is effective and reliable.
\begin{figure}
\centering
\includegraphics[width=4in]{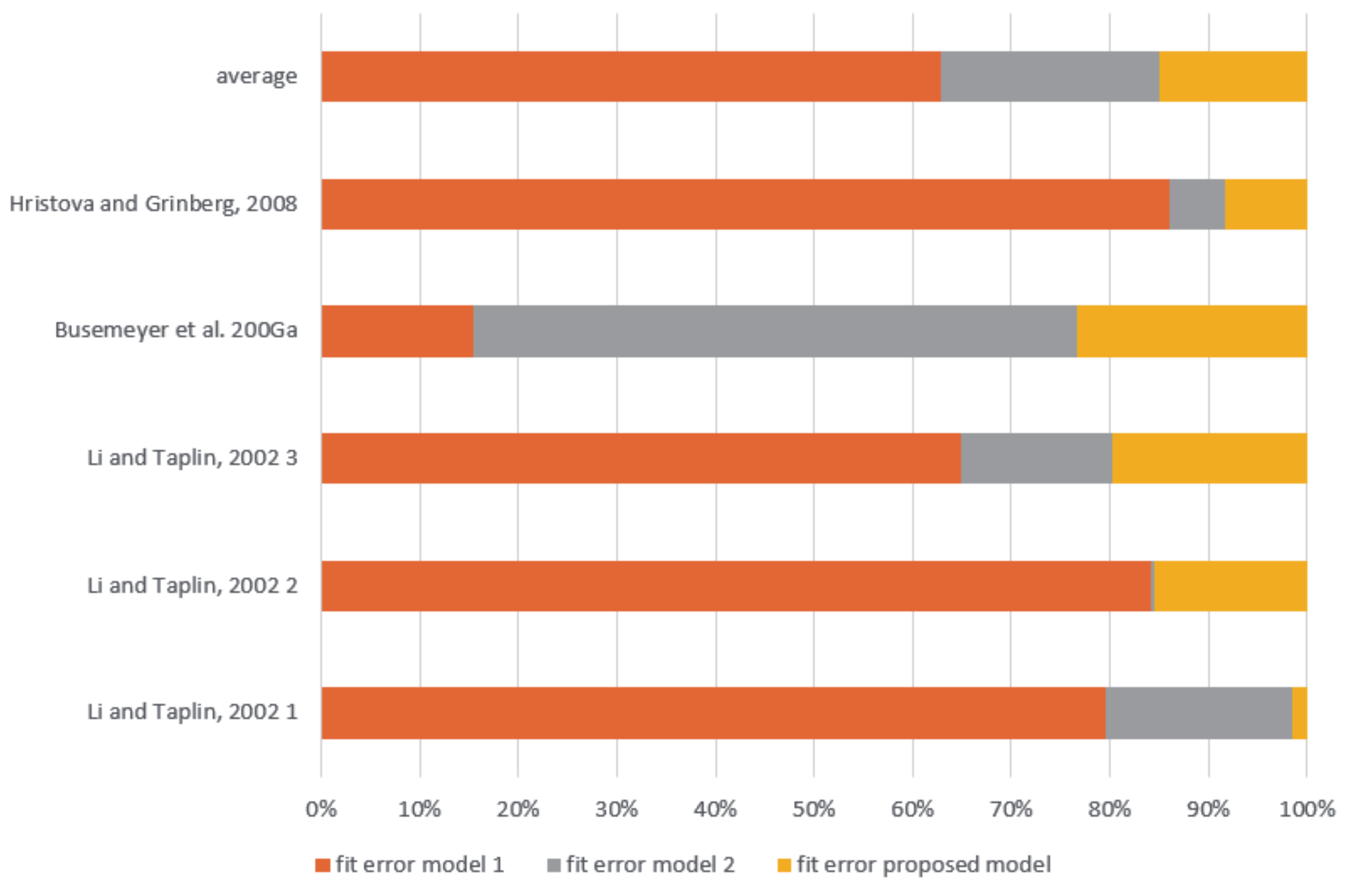}
\caption{Visualization of comparison results}
\label{model5}
\end{figure}
Fig.\ref{model5} visualizes the results from Table \ref{experimentresultsinliterature}, from which we can see that the result predicted by the proposed method is occupying the least area of the bar.

  The dilemma situation considered in this paper is prisoner's dilemma game with two strategies \cite{Tanimoto2007Relationship}. The dilemma strength \cite{Wang2015Universal} of the game discussed in this paper is $Dg'=Dr'=1$. There are also cases with dilemma strength different from $Dg'=Dr'=1$, depending on the payoff table to the players. Admittedly, the method proposed in this paper is designed to accommodate the paradoxical findings in dilemma strength $Dg'=Dr'=1$. Nevertheless, the method can well predict behaviours the player will take, as shown in Table \ref{comparisonmodel}. Comparing with other similar methods, the proposed Bayesian Network works with the least fit errors. In the prisoner's dilemma game, the prisoner will predict the other's action if he/she knows little about the rival. Therefore the probability of the prisoner to choose $Defect$ under unknown case will be smaller than the value computed from the classic way. The $Sure \ Thing \ Principle$ is violated because the belief in the prisoner's mind affects. On the other hand, the belief degree is not totally irregular. Lots of evidence have examined the value is closed within a small range. The advantages of Quantum-like Bayesian Network is it regards two strategies in people's mind as two wave functions, which will produce the interference effect. Hence, this paper proposes Belief Degree to represent the interference effect and utilizes Belief Distance to calculate the deviation from totally uncertainty. The belief entropy will produce a corresponding Belief Degree according to Belief Distance. We analyze the Prisoners' dilemma game with the model that applied our method and the prediction results are close to the observed probability with little fit error. In the end, we compare the model with two models which use a parameter called heuristic to predict the probability. The comparison results shows the effectiveness and reliability of our method.

\section{Acknowledgments}
The work is partially supported by National Natural Science Foundation of China (Grant No. 61671384), Natural Science Basic Research Plan in Shaanxi Province of China (Program No. 2016JM6018), Aviation Science Foundation (Program No. 20165553036).





\bibliographystyle{elsarticle-harv}

\bibliography{myreference}






\end{document}